%% file: iclr2025_conference.tex
\documentclass{article} 
\usepackage{iclr2025_conference,times}

\input{math_commands.tex}

\usepackage{hyperref}
\usepackage{url}
\usepackage{booktabs}
\usepackage{graphicx}
\usepackage{wrapfig}
\usepackage{lipsum}
\usepackage{derivative}
\title{TapWeight: Reweighting Pretraining Objectives for Task-Adaptive Pretraining}
\iclrfinaltrue

\author{Ruiyi Zhang , Sai Ashish Somayajula , Pengtao Xie\\
    UC San Diego\\
    {\tt\small \{ruz048, p1xie\}@ucsd.edu}
}

\begin{document}

\maketitle

\begin{abstract}
    Large-scale general domain pretraining followed by downstream-specific finetuning has become a predominant paradigm in machine learning. However, discrepancies between the pretraining and target domains can still lead to performance degradation in certain cases, underscoring the need for task-adaptive continued pretraining (TAP). TAP methods typically involve continued pretraining on task-specific unlabeled datasets or introducing additional unsupervised learning objectives to enhance model capabilities. While many TAP methods perform continued pretraining with multiple pretraining objectives, they often determine the tradeoff parameters between objectives manually, resulting in suboptimal outcomes and higher computational costs. In this paper, we propose TapWeight, a task-adaptive pretraining framework which automatically determines the optimal importance of each pretraining objective based on downstream feedback. TapWeight reweights each pretraining objective by solving a multi-level optimization problem. We applied TapWeight to both molecular property prediction and natural language understanding tasks, significantly surpassing baseline methods. Experimental results validate the effectiveness and generalizability of TapWeight. Our code is publicly available at \url{https://anonymous.4open.science/r/TapWeight-9A2E}.
\end{abstract}

\section{Introduction}

Foundation models pretrained on large-scale general domain corpora have achieved state-of-the-art performance across a wide range of tasks~\citep{He2021MaskedAA, devlin-etal-2019-bert, NEURIPS2020gpt3}. These models, which capture general knowledge for specific modalities such as text or images through unsupervised learning, are typically adapted to downstream tasks via finetuning. However, when there is a domain discrepancy between the pretraining corpus and the target task, direct finetuning of the pretrained model often fails to deliver optimal results~\citep{lee2020biobert,Chen2023MEDITRON70BSM,Xie2023EfficientCP}. To address this challenge, downstream task-adaptive continued pretraining, or task-adaptive pretraining (TAP), has been introduced. TAP bridges this gap by introducing an additional continued pretraining stage between general domain pretraining and task specific finetuning. For example, ~\citet{gururangan-etal-2020-dont} conducts task-adaptive pretraining by performing unsupervised learning on the unlabeled data of the downstream task. ~\citet{Wu2021DomainAdaptivePM} introduces an additional perturbation masking objective during continued pretraining of a BERT model~\citep{devlin-etal-2019-bert}, enhancing its performance on dialogue understanding tasks.

Among these, many existing task-adaptive pretraining methods consist of multiple pretraining objectives~\citep{Wu2021DomainAdaptivePM, gao2021simcse, Cui2023GroundedEA}, making it challenging to determine the relative importance of each objective. Some TAP methods assign equal weight to each pretraining objective~\citep{lee2020biobert,Wu2021DomainAdaptivePM}, disregarding their varying impact on downstream performance. For instance, ~\citet{gao2021simcse} shows that pretraining BERT with a contrastive learning (CL) objective results in better downstream performance on semantic textual similarity (STS) datasets than using masked language modeling (MLM) loss, indicating that the CL objective is more important than the MLM objective for these tasks. Other approaches attempt to manually tune the importance ratios through hyperparameter search~\citep{gao2021simcse}, which often results in suboptimal performance and increased computational costs. This issue becomes particularly severe when the number of pretraining objectives is large, such as with the task-adaptive pretraining of a popular molecular model Imagemol, which involves 5 distinct pretraining objectives~\citep{imagemol}.

\begin{figure}[t]
    \centering
    \vspace{-1cm}
    \includegraphics[width=\linewidth]{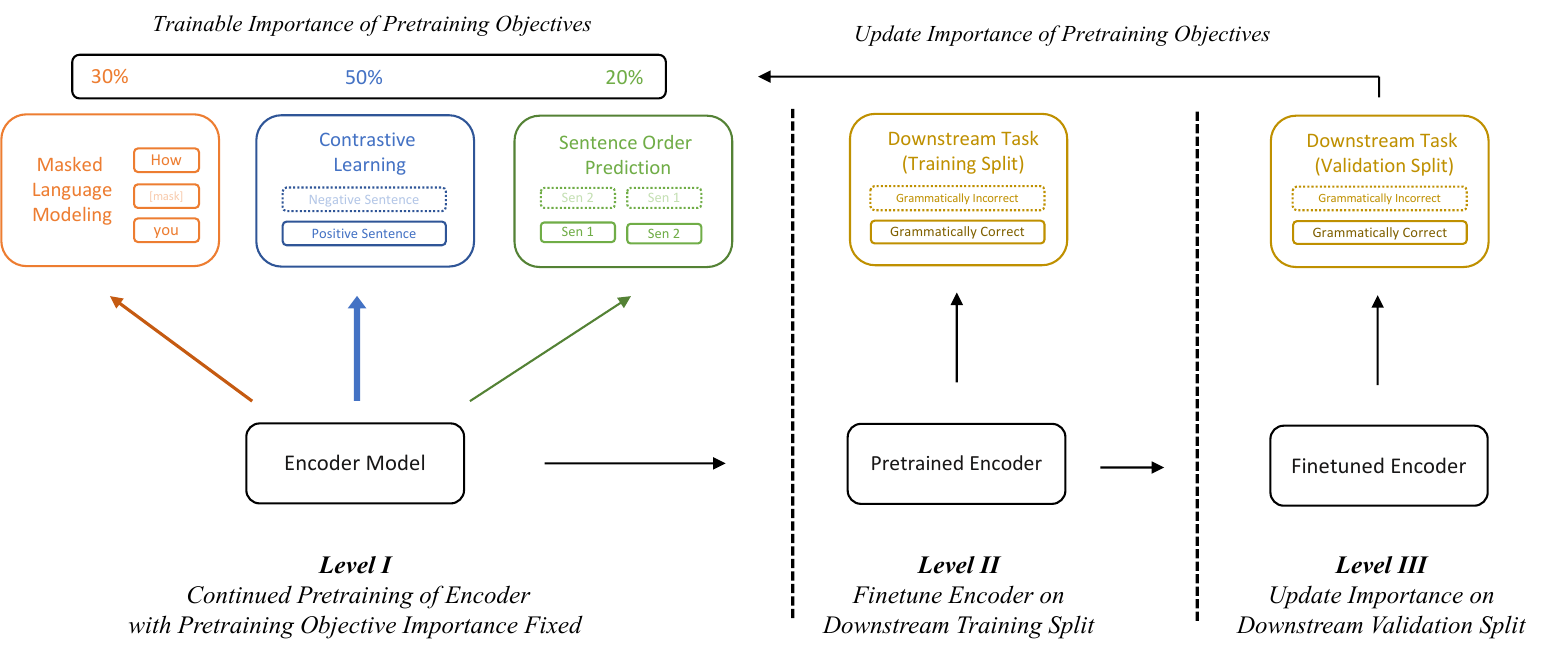}
    \caption{\textbf{An Overview of TapWeight.} In the first stage, the model undergoes multi-objective pretraining with fixed tradeoff ratios between objectives. In the second stage, the pretrained model is finetuned on the training split of the downstream dataset. In the third stage, the finetuned model is evaluated on the validation split of the downstream dataset to compute a loss, and the trainable tradeoff parameters fixed in the first stage are learned by minimizing this validation loss.}
    \label{fig:flowchart}
\end{figure}

To address the aforementioned challenges, we propose a novel framework, TapWeight, designed to learn the optimal tradeoff parameters between various pretraining objectives during task-adaptive pretraining. The goal is to learn these optimal tradeoff parameters such that the pretrained model, after finetuning on a downstream task, achieves the best downstream task performance. Our approach involves a three-level optimization framework to learn these parameters. In the first stage, we perform task-adaptive pretraining using initial tradeoff parameters, denoted as $\lambda$. These parameters are kept fixed during this stage and will be updated in subsequent stages. The resulting pretrained model is thus a function of $\lambda$. In the second stage, the pretrained model from above stage is finetuned on the training split of the downstream dataset. Consequently, the finetuned model becomes an implicit function of the tradeoff parameters. In the third stage, the finetuned model is evaluated on the validation split of the downstream dataset, and the tradeoff parameters $\lambda$ are optimized by minimizing the validation loss. This end-to-end process allows the three stages to dynamically influence one another, forming an integrated framework that optimizes task-adaptive pretraining process and enhances downstream task performance. Moreover, TapWeight is applicable to any pretrained model with multiple pretraining objectives, regardless of data modality or downstream task type, demonstrating superior generalizability compared to existing task-adaptive pretraining methods~\citep{Nishida2021,Cui2023GroundedEA}. Figure \ref{fig:flowchart} illustrates the complete framework of TapWeight.


We apply TapWeight for task-adaptive pretraining of both a molecule representation model, Imagemol~\citep{imagemol}, and a language model, RoBERTa~\citep{liu2019roberta}. Evaluating its performance across 13 molecular property prediction datasets and 8 natural language understanding tasks, TapWeight significantly outperforms baseline methods. The superior performance of TapWeight highlights its effectiveness and generalizability. Our contribution can be summarized as follows:

\begin{itemize}
    \item We propose TapWeight, an approach that automatically searches for the tradeoff parameters across multiple pretraining objectives and performs reweighted task-adaptive pretraining. TapWeight is formulated within a multi-level optimization (MLO) framework. We employ an efficient gradient descent algorithm to solve the MLO problem, obtaining the optimal tradeoff parameters for multiple pretraining objectives. Our implementation of TapWeight is publicly available.
    \item We apply TapWeight for task-adaptive pretraining of a molecule representation model and a language model. Extensive experiments on 13 downstream datasets in molecular property prediction and 8 datasets in natural language understanding underscore its effectiveness and generalizability.
\end{itemize}

\section{Related Works}
\subsection{Domain / Task Adaptive Pretraining}

To bridge the gap between general domain pretraining and downstream tasks in a specific domain, domain-adaptive pretraining (DAP) and task-adaptive pretraining (TAP) have been introduced~\citep{gururangan-etal-2020-dont}. DAP performs continued pretraining on a large, unlabeled corpus from a similar domain as the downstream task. For example, BioBERT continues to pretrain a BERT model on a large-scale biomedical corpus, enhancing its performance on a variety of biomedical text mining tasks~\citep{lee2020biobert}. Similarly, LegalBERT continues to pretrain a BERT model on legal documents to improve performance on legal NLP tasks~\citep{chalkidis-etal-2020-legal}, while SciBERT leverages a large multi-domain corpus of scientific publications for further pretraining, enhancing its effectiveness on scientific NLP tasks~\citep{beltagy-etal-2019-scibert}. More recently, MEDITRON performs continued pretraining of a Llama-2 model with 80 billion parameters on text in medical domain, showing significant performance gains on major medical benchmarks~\citep{Chen2023MEDITRON70BSM}. 

Although DAP significantly improves model performance on downstream tasks, it needs a large corpus of unlabeled data in a specific domain, which is not always available. To address this limitation, multiple task-adaptive pretraining (TAP) methods have emerged, which do not rely on additional domain-specific corpora beyond the downstream dataset itself. TAP methods can also be viewed as a novel finetuning process, where standard finetuning is preceded by low-cost continued pretraining. For instance, TAPT performs continued pretraining directly on the unlabeled training split of the downstream dataset~\citep{gururangan-etal-2020-dont}. TAPTER first trains new word embeddings using the unlabeled training split of the downstream dataset, and then use these embeddings for continued pretraining of the model~\citep{Nishida2021}. SimCSE introduces an additional constrastive learning loss in addition to the original masked language modelling loss to further pretrain a RoBERTa model, specifically enhancing its capability on standard semantic textual similarity tasks \citep{gao2021simcse}. While existing TAP methods are effective, they are typically tailored to specific downstream tasks or data modalities~\citep{Wu2021DomainAdaptivePM,Cui2023GroundedEA}. In contrast, TapWeight is applicable to any pretrained model with multiple pretraining objectives, underscoring its broad generalizability. 

\subsection{Multi-level Optimization}
Many machine learning tasks can be formulated as multi-level optimization (MLO) problems, such as neural architecture search~\citep{liu2018darts,chen2019progressive,xu2020pcdarts}, meta learning~\citep{maml,imaml,zhang-etal-2024-autolora}, and hyperparameter optimization~\citep{pmlr-v108-lorraine20a, lorraine2018stochastic, mackay2018selftuning}. MLO problems consist of multiple levels of optimization problems that are mutually dependent, making it challenging for common automatic differentiation algorithms to handle them. To tackle this challenge, multiple algorithms~\citep{pmlr-v108-lorraine20a, liu2018darts, imaml} and libraries~\citep{choe2023betty, choe2023making} have been proposed to efficiently compute gradients in MLO problems.

Recently, MLO techniques have been widely adopted in data reweighting and task reweighting. In these methods, the weights of data or tasks are often treated as hyperparameters and optimized in the upper levels of MLO problems. For example, MetaWeightNet learns an explicit weighting function for each data point to maximize the performance on a small amount of unbiased meta-data~\citep{han2018coteaching}. DoGE optimizes weights for each data domain using a small proxy model to guide the pretraining of larger models~\citep{fan2024doge}. MetaWeighting learns tradeoff parameters for each task in multi-task learning to minimize generalization loss~\citep{mao-etal-2022-metaweighting}.  Our method also falls within this category, with a specific focus on reweighting pretraining objectives for downstream task-adaptive continued pretraining.

\section{Method}

\subsection{Overview}

Given $n$ continued pretraining objectives $\mathcal{T}_1, \mathcal{T}_2, ... \mathcal{T}_n$ and their corresponding training losses $\mathcal{L}_1, \mathcal{L}_2, ... \mathcal{L}_n$, we formulate the multi-objective continued pretraining loss $\mathcal{L}_{pt}$ as:

\begin{equation}
    \mathcal{L}_{pt}(\theta, \lambda, \mathcal{D}_{pt})=\sum_{i=1}^n\lambda_i\mathcal{L}_i(\theta, \mathcal{D}_{pt})
\end{equation}

where $\mathcal{D}_{pt}$ is the unsupervised pretraining dataset, $\theta$ denotes the pretraining model parameters, and $\lambda_i$ is the tradeoff parameter for each pretraining objective. We denote the target downstream task as  $\mathcal{D}_{ft}$ and split it into $\mathcal{D}_{tr}, \mathcal{D}_{val}$ and $ \mathcal{D}_{ts}$, which are training, validation and test splits respectively. 

In our framework, TapWeight, we aim to automatically search for the optimal tradeoff weights $\lambda=\{\lambda_1,...,\lambda_n\}$, so that the pretrained model will achieve highest performance on $\mathcal{D}_{val}$ after finetuned on a downstream dataset $\mathcal{D}_{ft}$. To achieve this, our method consists of three end-to-end stages. In the first stage, we perform continued pretraining of the model, with tradeoff weights tentatively fixed. In the second stage, we conduct finetuning of the pretrained model on the training split of the downstream dataset. In the third stage, we compute a loss by applying the finetuned model on the validation split of the downstream dataset, and optimize the tradeoff parameters by minimizing this loss. We next formally define these three stages under a multi-level optimization framework.

\subsection{TapWeight Framework}


\paragraph{Level I} In the first level, we aim to perform continued pretraining for the model. Formally, the optimization problem (OP) is to optimize the model weights $\theta$ to minimize the multi-objective pretraining loss $\mathcal{L}_{pt}$ on a unlabeled dataset $\mathcal{D}_{pt}$:

\begin{align}
    \theta^*(\lambda)=\argmin_\theta\mathcal{L}_{pt}(\theta, \lambda, \mathcal{D}_{pt})
\end{align}

Since the optimal solution $\theta^*$ to this problem depends on the value of the tradeoff parameter, it is an implicit function of $\lambda$, denoted as $\theta^*(\lambda)$.

\paragraph{Level II} In the second level, we aim to finetune the pretrained model with optimal parameters $\theta^*$ obtained from previous level on the downstream dataset. However, formulating the optimization problem here with the same set of parameters $\theta$ as the lower level imposes high computation and memory burdens, as it requires differentiating through the whole gradient update trajectory in the lower level~\citep{imaml}. Optimizing distinct sets of parameters at different levels enables the use of implicit differentiation methods, which significantly reduces computational costs, as detailed in Section \ref{sec:optim}. Therefore, we create a model with new parameters $\omega$ that are different from those in the pretrained model, but with a regularization loss $\mathcal{R}$ between $\omega$ and $\theta$ to encourage them to be close. This proximal constraint casts strong dependence between $\omega^*$ and $\theta^*$, closely resembling the real finetuning process. Formally, the OP in this level is to optimize $\omega$ by minimizing the weighted summation of finetuning loss $\mathcal{L}_{tr}$ and the proximal regularization loss $\mathcal{R}$:

\begin{align}
\label{eq:level2}
    \omega^*(\theta^*(\lambda))&=\argmin_{\omega}\mathcal{L}_{tr}(\omega,\mathcal{D}_{tr})+\gamma\mathcal{R}(\omega,\theta^*(\lambda))
\end{align}

where $\mathcal{D}_{tr}$ is the training split of the downstream dataset, and $\gamma$ is a tradeoff hyperparameter to balance the finetuning loss and regularization loss. In practice, we select the mean squared error (MSE) loss as the regularization loss $\mathcal{R}$. The optimal solution of $\omega$ in this level is a function of $\theta^*$ due to the loss term $\mathcal{R}$, which is in turn a function of $\lambda$, denoted as $\omega^*(\theta^*(\lambda))$.

\paragraph{Level III}
In the third level, we aim to search for the optimal tradeoff parameters $\lambda^*$ between pretraining objectives. Formally, the OP in this level is to optimize $\lambda$ to minimize the validation loss $\mathcal{L}_{val}$:

\begin{align}
    & \min_\lambda\mathcal{L}_{val}(\omega^*(\lambda),\mathcal{D}_{val})
\end{align}

where $\mathcal{D}_{val}$ is the validation split of the downstream dataset. 

\paragraph{Multi-level Optimization Framework} In this way, we formulate a three-level optimization problem with OPs in different levels mutually dependent on each other:

\begin{align}
\label{eq:mlo}
    & \min_\lambda\mathcal{L}_{val}(\omega^*(\lambda),\mathcal{D}_{val})\\
    s.t. \quad \omega^*(\theta^*(\lambda))&=\argmin_{\omega}\mathcal{L}_{tr}(\omega,\mathcal{D}_{tr})+\gamma\mathcal{R}(\omega,\theta^*(\lambda)) \notag\\ 
     \theta^*(\lambda)&=\argmin_\theta\mathcal{L}_{pt}(\theta, \lambda, \mathcal{D}_{pt}) \notag
\end{align}

By solving this multi-level optimization problem, we are able to reweight each continued pretraining objective based on feedback from validation performance on downstream tasks. In practice, both $\theta$ and $\omega$ in Equation \ref{eq:mlo} are initialized with model weights from general-domain pretraining.

\subsection{Optimization Algorithm}
\label{sec:optim}
In this section, we illustrate the algorithm we use to efficiently approximate the gradient of loss $\mathcal{L}_{val}$ in the third level with respect to the tradeoff parameter $\lambda$. This full derivative $\frac{d\mathcal{L}_{val}}{d\lambda}$ can be computed with the following equation using chain rule:

\begin{align}
\label{eq: grad}
    \frac{d\mathcal{L}_{val}}{d\lambda} ={\color{olive}\frac{\partial \mathcal{L}_{val}}{\partial \omega^*}} \times {\color{red}\frac{\partial \omega^*}{\partial \theta^*}} \times {\color{red} \frac{\partial \theta^*}{\partial\lambda}}
\end{align}

In the right hand side of Equation \ref{eq: grad}, the {\color{olive} olive} term, a partial derivative vector, can be directly computed with popular automatic differentiation libraries, such as Pytorch~\citep{pytorch}. However, directly computing the two {\color{red} red} terms, which are best-response Jacobian matrices, can be computationally prohibitive due to the lack of analytical solutions to these optimization problems. Inspired by previous works~\citep{pmlr-v108-lorraine20a, pmlr-v139-idarts}, we use Implicit Function Theorem (IFT) based methods to approximate the best-response Jacobian matrices. We include more details of IFT based gradient computation method in Appendix \ref{appen:alg}. In this way, we are able to compute both  {\color{red} red} terms in Equation \ref{eq: grad} efficiently, thereby obtaining the gradient of $\mathcal{L}_{val}$ with respect to $\lambda$. We then optimize the tradeoff parameter $\lambda$ with gradient descent. The complete algorithm is implemented using the Betty library~\citep{choe2023betty,choe2023sama}.

\begin{table}[t]
    \centering
    \setlength{\tabcolsep}{4pt}
    \begin{tabular}{c|cccccccc|c}
    \toprule
    \bf Method     &\bf BACE  &\bf BBBP  &\bf ClinTox&\bf Sider  &\bf Tox21  &\bf ToxCast  &\bf HIV  &\bf MUV  &\bf Avg.   \\
    \texttt{\small Dataset Size} & \texttt{\scriptsize 1,513} &\texttt{\scriptsize2,039}&\texttt{\scriptsize1,478}&\texttt{\scriptsize1,427}&\texttt{\scriptsize7,831}&\texttt{\scriptsize8,575}&\texttt{\scriptsize41,127}&\texttt{\scriptsize93,087}& \\
    \midrule
    \small AttrMask     &77.2	&70.2&	68.6&	60.4&	74.2&	62.5&	74.3&	73.9&	70.2   \\
    \small ContextPred     &  78.6&	\textbf{71.2}&	73.7&	59.3&	73.3&	62.8&	75.8&	72.5&	70.9   \\
    \small GraphMVP     & 76.8&	68.5&	79.0&	62.3&	74.5	&62.7&	74.8&	75.0&	71.7  \\
    \small Imagemol     &80.1&	67.3&	78.5&	63.6&	76.5&	65.4&	75.6&	78.4&	73.2   \\
    \small \textbf{TapWeight (ours)}     &  \textbf{83.1}&	\textbf{71.2}&\textbf{81.3}	&\textbf{64.5}	&\textbf{77.0}	&\textbf{66.1}	&	\textbf{78.4}&\textbf{80.5}	&\textbf{75.3}	\\
    \bottomrule
    \end{tabular}
    \caption{Results of molecular property prediction on 8 classification tasks in MoleculeNet benchmark, in terms of AUROC. The best results are shown in \textbf{bold}.}
    \label{tab:clf_mol}
\end{table}

\section{Experiments}

\subsection{Molecular Property Prediction}

In this section, we use TapWeight for task-adaptive pretraining of molecular image models and validate the effectiveness of our framework on the downstream task of molecular property prediction.

\subsubsection{Preliminary} 

Given a large unlabeled molecular dataset $\mathcal{D}=\{x_i\}_{1\leq i \leq n}$ containing millions of molecules, we define a multi-objective continued pretraining loss inspired by Imagemol~\citep{imagemol}:

\begin{align}
    \mathcal{L}(x)=\lambda_1\mathcal{L}_{mg1}(x)+\lambda_2\mathcal{L}_{mg2}(x)+\lambda_3\mathcal{L}_{mg3}(x)+\lambda_4\mathcal{L}_{jpp}(x)+\lambda_5\mathcal{L}_{mcl}(x)
\end{align}

where $x$ represents a molecular image, and $\lambda=\{\lambda_i\}_{1\leq i \leq 5}$ are tradeoff parameters. $\mathcal{L}_{mg1}$, $\mathcal{L}_{mg2}$, and $\mathcal{L}_{mg3}$ are MACCS key \citep{Durant2002MACCS} clustering-based classification losses with different number of clusters. $\mathcal{L}_{jpp}$ is a jigsaw puzzle prediction loss, where the model solves a jigsaw puzzle on the same molecular image. $\mathcal{L}_{mcl}$ is a mask-based contrastive learning loss, which generates constrastive pairs by masking molecular images. Details of these pretraining objectives can be found in Appendix \ref{appen:molecule_task}. 

The multi-objective loss $\mathcal{L}$ is optimized on the complete unlabeled dataset $\mathcal{D}$ to train a molecular image encoder. The learnt encoder can be further finetuned on downstream datasets for various molecular tasks. Existing approaches typically set the tradeoff parameters $\lambda$ equally across different pretraining objectives, overlooking the varying contributions of each objective to specific downstream tasks~\citep{imagemol}. We address this challenge by applying TapWeight framework for continued pretraining of the molecular image encoder. 

\subsubsection{Experimental Settings} 
\label{sec:mol_exp_set}
We perform continued pretraining of a pretrained Imagemol model on a dataset $\mathcal{D}$, consisting of 1 million molecules from PubChem~\citep{kim2023pubchem}. For downstream tasks, we employ the MoleculeNet benchmark, which includes 8 classification datasets focused on predicting biophysical and physiological properties essential for drug discovery~\citep{Wu2017MoleculeNetAB}. We generate the training, validation and test split of these downstream datasets by applying scaffold splitting with an 8:1:1 ratio. We use AUROC as the evaluation metric for all classification datasets, MAE for Qm7 and Qm9 datasets, and RMSE for all other regression datasets. In addition to Imagemol, we benchmark against Graph Neural Network (GNN)-based molecular property prediction methods, including pretraining approaches such as attribute masking, context prediction~\cite{hu2020pretraining}, and GraphMVP~\citep{liu2022graphmvp}. The pretrained molecular image encoder is based on a ResNet18 model, with the final classification layer removed~\citep{He2015resnet}. We set the number of clusters for the loss terms $\mathcal{L}_{mg1}$, $\mathcal{L}_{mg2}$, and $\mathcal{L}_{mg3}$ to 100, 1,000, and 10,000, respectively. During the continued pretraining, we set the unrolling step in the MLO framework to be 1. We use the SGD optimizer with a step learning rate scheduler across all three optimization levels. All experiments are conducted on NVIDIA A100 GPUs. More detailed descriptions are provided in the Appendix for the datasets (\ref{appen:molecule_data}), baselines (\ref{appen:molecule_bsl}), and hyperparameter settings (\ref{appen:molecule_hyper}).  

\subsubsection{Results}

\begin{wraptable}{r}{8cm}
    \centering
    \vspace{-0.5cm}

    \setlength{\tabcolsep}{3pt}
    \vspace{0.2cm}
    \begin{tabular}{c|ccccc }\toprule
        \bf Method & \bf Freesolv	&\bf Esol	&\bf Lipo	&\bf Qm7	&\bf Qm9         \\ 
        \texttt{\footnotesize Dataset Size} & \texttt{\scriptsize 642} &\texttt{\scriptsize 1,128}&\texttt{\scriptsize4,200}&\texttt{\scriptsize6,830}&\texttt{\scriptsize133,885}\\
        \midrule
        \small AttrMask   & 2.95 & 1.37 & 0.81 &161.7 &5.03 \\
        \small ContextPred  & 3.01 & 1.35 & 0.83 &153.2 &4.95 \\
        \small GraphMVP   & 2.21 & 1.13 & 0.79 &134.5 &4.76  \\
        \small Imagemol   & 3.04&	1.11&	\bf 0.76&	141.0&	4.52\\
        \small \textbf{TapWeight (ours)} &\bf 1.91	&\bf1.06&\bf	0.76&\bf	126.0&	\bf 4.28 \\ \bottomrule
    \end{tabular}
        \caption{Results of molecular property prediction on 5 regression tasks in MoleculeNet benchmark. The best results are shown in \textbf{bold}.}
    \vspace{-0.2cm}
    \label{tab:reg_mol}
\end{wraptable}

Table \ref{tab:clf_mol} show the results of various methods across 8 molecular property classification tasks from MoleculeNet benchmark. Our method outperforms all baseline methods on all 8 datasets, showcasing the effectiveness of our method. On average, our method achieves an AUROC of 75.3, compared to 73.2 for the Imagemol model without continued pretraining.  Similarly, Table \ref{tab:reg_mol} displays the results for 5 regression tasks in the MoleculeNet benchmark, where our method once again surpasses all baselines on each task. Experimental results validate the effectiveness of our method on both classification and regression tasks. Specifically, the superior performance of our method over Imagemol validates the necessity of downstream-guided continued pretraining following general pretraining. Notably, our method consistently outperforms baseline approaches regardless of the size of the finetuning dataset, demonstrating the robustness of our approach. It is worth mentioning that TAPT~\citep{gururangan-etal-2020-dont} is not applicable to this task, as clustering-based losses, such as $\mathcal{L}_{mg3}$, are not well-suited for direct application on small unlabeled datasets where the number of data points is smaller than the predefined number of clusters. In contrast, TapWeight does not face such limitations, demonstrating its generalizability.

\subsection{Natural Language Understanding}

In this section, we validate the effectiveness of TapWeight for continued pretraining of a masked language model (MLM) with its application to natural language understanding tasks.

\subsubsection{Prelimimary} 

Given a large-scale raw-text dataset $\mathcal{D}=\{x_i\}_{1\leq i \leq n}$ consisting of millions of sentences, we define the following continued pretraining loss:

\begin{align}
    \mathcal{L}(x)=\lambda_1\mathcal{L}_{mlm}(x)+\lambda_2\mathcal{L}_{cl}(x)+\lambda_3\mathcal{L}_{sop}(x)
\end{align}

where $x$ is a sentence, and $\lambda=\{\lambda_i\}_{1\leq i \leq 3}$ are tradeoff parameters. $\mathcal{L}_{mlm}$ represents the masked language model loss, which involves randomly masking tokens in the input sentences and predicting these masked tokens~\citep{devlin-etal-2019-bert}. $\mathcal{L}_{cl}$ denotes the contrastive learning loss, where an input sentence is used to predict itself with standard dropout applied as noise~\citep{gao2021simcse}. $\mathcal{L}_{sop}$ is the sequence ordering prediction loss, which emphasizes inter-sentence conherence~\citep{Lan2020ALBERT:}. We include details of these losses in Appendix \ref{appen:nlu_task}. 

The multi-objective loss $\mathcal{L}$ is optimized on the raw text dataset $\mathcal{D}$ for continued pretraining of a Transformer encoder. The learnt encoder can then be finetuned on downstream NLP datasets. In existing works~\citep{gao2021simcse}, the tradeoff parameters $\lambda$ for different pretraining objectives require manual hyperparameter tuning, which is time-consuming and often leads to suboptimal results. We address this challenge by applying TapWeight for the continued pretraining of a Transformer encoder, enabling the automatic determination of the importance for each objective.

\begin{table}[t]
    \centering
    \setlength{\tabcolsep}{4pt}
    \begin{tabular}{c|cccccccc|c}
    \toprule
    \bf Method     &\bf MNLI&	\bf QNLI &\bf 	QQP 	&\bf RTE &\bf 	SST 	&\bf MRPC 	&\bf CoLA &\bf STSB  &\bf Avg.   \\
    \texttt{\small Dataset Size} & \texttt{\scriptsize 392,702} &\texttt{\scriptsize 104,743}&\texttt{\scriptsize 363,871}&\texttt{\scriptsize 2,490}&\texttt{\scriptsize 67,349}&\texttt{\scriptsize 3,668}&\texttt{\scriptsize 8,551}&\texttt{\scriptsize 5,749}& \\
    \midrule

    \small Finetuning     & 86.7	&\bf 92.8&	90.3&77.8&94.8&89.3&61.6	&\bf 91.2&85.6  \\
    \small SimCSE     &85.6&	90.1&	90.7&	74.6&	91.1&	89.2&	59.7&	91.0&	83.6   \\
    \small TAPT     & 85.2	& 91.3&	90.2&78.2&93.7&90.1&61.5	&90.9&85.1  \\
    \small \textbf{TapWeight (ours)}     & \bf 86.8&	92.5&	\bf 91.1&	\bf 80.7	&\bf 94.9&	\bf 90.2&\bf 62.3&\bf 91.2&	\bf 86.2	\\
    \bottomrule
    \end{tabular}
    \caption{Results of different methods in GLUE benchmark. All methods are applied to a RoBERTa-base model. The best results are shown in \textbf{bold}.}
    \label{tab:glue}
\end{table}

\subsubsection{Experimental Settings}
\label{sec:nlu_exp_set}
We perform continued training of a pretrained RoBERTa-base model on a raw-text dataset $\mathcal{D}$ consisting of 1 million sentences from Wikipedia~\citep{gao2021simcse}. For downstream evaluation, we use the GLUE benchmark, which comprises 8 natural language understanding tasks, including sentiment analysis, semantic similarity prediction, and grammaticality classification~\citep{wang2018glue}.  Following standard practices, we use the original GLUE development set as the test set in our experiments, and randomly split the original training set into a training set and validation set with a ratio of 8:1. We use Matthew's Correlation for the CoLA dataset, Pearson/Spearman Correlation for the STS-B dataset, and accuracy for all other datasets. Our baseline methods are all based on a RoBERTa-base model, including direct finetuning, TAPT based continued pretraining and SimCSE based continued pretraining. When applying TapWeight on the RoBERTa-base encoder, we set the unrolling step in the MLO framework to 1. We use an Adam optimizer with a step learning rate scheduler across all three optimization levels. All experiments are conducted on NVIDIA A100 GPUs. More detailed descriptions are provided in the Appendix for the datasets (\ref{appen:nlu_data}), baselines (\ref{appen:nlu_bsl}), and hyperparameter settings (\ref{appen:nlu_hyper}).

\subsubsection{Results }

Table \ref{tab:glue} presents the results of various methods on 8 natural language understanding tasks from the GLUE benchmark. TapWeight consistently outperforms all baseline methods across all 8 datasets, showcasing the effectiveness of our method. On average, our method achieved a score of 86.2, while finetuning a RoBERTa model without continued pretraining only got 85.6. The superior performance on both molecule property prediction and natural language understanding highlights the generalizability of our method across multiple data modalities and downstream tasks. Moreover, our method surpasses the SimCSE method on all 8 tasks, showcasing the effectiveness of reweighting pretraining objectives, as SimCSE uses a fixed ratio between MLM and CL losses during continued pretraining. Additionally, TapWeight outperforms the RoBERTa+TAPT approach, demonstrating that our strategy of leveraging downstream datasets by reweighting pretraining objectives is more effective than simply pretraining the model with unlabeled downstream data, as TAPT does.

\subsection{Ablation Studies}

\begin{table}
    \centering
    \setlength{\tabcolsep}{3.5pt}
    \begin{tabular}{c|cccccccc|c}
    \toprule
    \bf Method     &\bf BACE  &\bf BBBP  &\bf ClinTox&\bf Sider  &\bf Tox21  &\bf ToxCast  &\bf HIV  &\bf MUV  &\bf Avg.   \\
    \midrule

    \small CP w/o Reweighting     & 78.8&	66.1&	77.4&	60.3&	74.6	&62.7&	76.9&	71.6&	71.1  \\
    \small TapWeight w/o MLO     &83.0&	68.5&	79.5&	63.5&	76.3&	65.9&	77.2&	77.3&	73.9   \\
    \small \textbf{TapWeight}     &  \textbf{83.1}&	\textbf{71.2}&\textbf{81.3}	&\textbf{64.5}	&\textbf{77.0}	&\textbf{66.1}	&	\textbf{78.4}&\textbf{80.5}	&\textbf{75.3}	\\
    \bottomrule
    \end{tabular}
    \caption{\textbf{Ablation Studies.} Results of molecular property classification using our method and baseline methods, in terms of AUROC. The best results are shown in \textbf{bold}.}
    \label{tab:abl}
\end{table}

In this section, we perform ablation studies to evaluate the effectiveness of individual components within our framework. All experiments are conducted on the classification tasks in the molecular property prediction benchmark.

\paragraph{Pretraining Objective Reweighting} We validate the effectiveness of our pretraining objective reweighting strategy by comparing our method to continued pretraining with a fixed importance for each objective. As shown in Table \ref{tab:abl}, our method outperforms this baseline (CP w/o Reweighting) across all datasets, demonstrating the advantage of dynamically reweighting pretraining objectives in the continued pretraining process.

\paragraph{Multi-level Optimization} We validate the effectiveness of the multi-level (tri-level) optimization (MLO) framework by reducing our method to a bi-level optimization (BLO)~\citep{skillearn} based method. Specifically, we merge the first and second level of problems from the TapWeight framework to form the lower-level problem in the new BLO baseline, where the model is optimized jointly using both the unsupervised pretraining loss on the unlabeled continued pretraining dataset $\mathcal{D}_{pt}$ and the finetuning loss on the training split of the downstream dataset $\mathcal{D}_{tr}$. In the upper-level problem, the importance for each pretraining objective is learned using the validation split of the downstream dataset. Formally, we define the following BLO problem:

\begin{align}
    & \min_\lambda\mathcal{L}_{val}(\theta^*(\lambda),\mathcal{D}_{val})\\
     s.t. \quad \theta^*(\lambda)&=\argmin_\theta\mathcal{L}_{pt}(\theta, \lambda, \mathcal{D}_{pt})+\gamma\mathcal{L}_{tr} (\theta,\mathcal{D}_{tr})\notag
\end{align}

However, optimizing these two types of losses in the lower level requires extensive tuning of the tradeoff parameters $\gamma$, and often leads to competition between losses which results in performance decrease.  As shown in Table \ref{tab:abl}, our MLO based reweighting method outperforms the BLO based approach across all datasets, highlighting the advantage of formulating multiple optimization problems. Nevertheless, BLO method still outperforms the baseline continued pretraining methods with fixed tradeoff parameters, indicating the necessity of using reweighting strategies. 

\begin{figure}
    \vspace{-0.5cm}
    \centering
    \includegraphics[width=\linewidth]{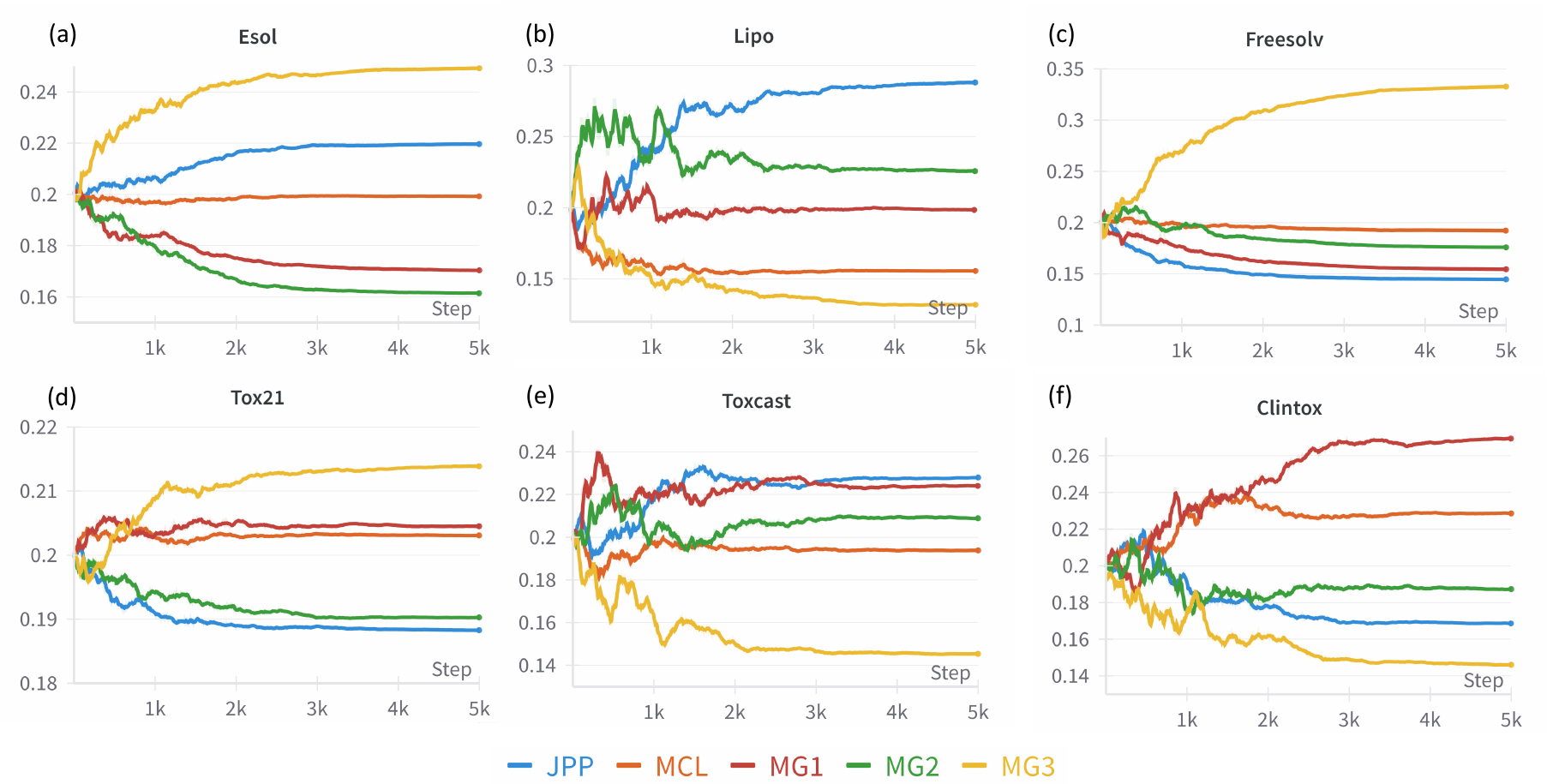}
    \caption{ Evolution of the tradeoff parameter $\lambda$ over the training steps of TapWeight on the following downstream datasets: (a) Esol, (b) Lipo, (c) Freesolv, (d) Tox21, (e) Toxcast, and (f) Clintox. }
    \label{fig:trend}
\end{figure}

\subsection{Qualititive Analysis}

In this section, we present the evolution trend of the pretraining objective weights along the training trajectory using our method. As shown in Figure \ref{fig:trend}, we plot the value of $\lambda$ for 3 regression tasks (Esol, Lipo, Freesolv) and 3 classification tasks (Tox21, Toxcast, Clintox) with respect to the global training step. Our observations reveal that different downstream datasets require varying importance for each pretraining objective. For example, the JPP pretraining objective, $\mathcal{L}_{jpp}$, plays an key role in Lipo and Toxcast datasets, whereas the MG3 pretraining objective, $\mathcal{L}_{mg3}$, is more critical for Esol, Freesolv and Tox21 datasets. The diverse requirements of pretraining objectives across downstream datasets emphasize the need for a reweighting method like TapWeight, providing a clear explanation for why our method outperforms baseline approaches.


\subsection{Computation Cost}

\begin{wraptable}{r}{6cm}
    \centering
    \vspace{-0.6cm}
    
    \setlength{\tabcolsep}{3.5pt}
    \begin{tabular}{c|ccc }\toprule
        \bf Dataset & \bf \small FT	&\bf \small CP+FT	&\bf \small TapWeight     \\ 

        \midrule
        MUV   & $\times 1$ & $\times 2.18$ & $\times 3.29$  \\
        Qm9  & $\times 1$ & $\times 2.76$ & $\times 3.93$  \\
        QQP & $\times 1$ & $\times 2.54$ & $\times 3.76$ \\
        \bottomrule
    \end{tabular}
    \caption{Training cost of baseline methods and our method TapWeight.}
    \vspace{-0.2cm}
    \label{tab:cost}
\end{wraptable}

In this section, we compare the training time of our method with baseline methods on the QQP, MUV and Qm9 datasets, as shown in Table \ref{tab:cost}. We use finetuning (FT) and continued pretraining with a fixed tradeoff ratio (CP+FT) as baselines,  normalizing the time cost of FT as 1.  While TapWeight results in an increase in training time compared to FT and CP+FT, its substantial improvement across multiple downstream tasks generally justifies the additional cost. However, in real-world applications where training time is a critical factor, TapWeight may not be the ideal choice, representing a limitation of our approach.

\section{Conclusion and Future Work}


In this paper, we propose a task-adaptive continued pretraining method that dynamically reweights each pretraining objective within a multi-level optimization framework. Unlike previous approaches that use a fixed ratio between pretraining objectives, our method adjusts the importance of each objective based on feedback from downstream datasets. Experiments in both molecule property prediction and natural language understanding validate the effectiveness and generalizability of our method. 

Given the success of TapWeight, several promising future research directions emerge. For instance, large multimodal pretrained models have recently gained popularity~\citep{liu2023llava, zhu2024minigpt}. The combination of multiple modalities introduces a greater number of potential continued pretraining objectives, presenting necessities of applying TapWeight in this context. Additionally, exploring objective reweighting strategies for general pretraining, rather than task-adaptive pretraining (TAP), is another promising direction. Unlike TAP, pretraining objective reweighting in general domain presents greater challenges for algorithm efficiency, as general pretraining typically incurs significantly higher computational costs. 

\section{Reproducibility Statement}

We provide the code of TapWeight at \url{https://anonymous.4open.science/r/TapWeight-9A2E}. In the code repo, we provide instructions on how to reproduce experimental results for both molecule property prediction and natural language understanding. Furthermore, we include detailed experimental settings of molecule property prediction in Section \ref{sec:mol_exp_set}, with more information on selection of hyperparameters in Appendix \ref{appen:molecule_hyper}. For natural language understanding, we include detailed experimental settings in Section \ref{sec:nlu_exp_set}, with more information on selection of hyperparameters in Appendix \ref{appen:nlu_hyper}.

\bibliography{iclr2025_conference}
\bibliographystyle{iclr2025_conference}

\clearpage

\appendix

\section{Optimization Algorithm}
\label{appen:alg}
In this section, we give an example to briefly illustrate how to use Implicit Function Theorem (IFT) to compute best-response Jacobian matrices. Take $\pdv{\theta^*}{\lambda}$ term in Equation \ref{eq: grad} as an example: although $\theta^*$ is an implicit function of $\lambda$, the exact value of $\theta^*(\lambda)$ given a value of $\lambda$ is usually approximated with gradient descent algorithms. As there is no analytical solution of $\theta^*(\lambda)$, it is difficult to directly compute the gradient $\pdv{\theta^*}{\lambda}$.  To tackle this challenge, we compute this gradient using IFT following previous literature~\citep{pmlr-v108-lorraine20a}:

\begin{align}
\label{eq:ift}
    { \pdv{\theta^*}{\lambda}} = {- \color{red}[\nabla^2\mathcal{L}_{pt}(\theta)]^{-1}}  \times {\color{olive}\pdv{\mathcal{L}_{pt}}{\theta, \lambda^T}}
\end{align}

The {\color{olive} olive} term, a second-order mixed partial derivative matrix, can be directly computed using automatic differentiation. Nevertheless, directly computing the {\color{red} red} term, which is the invert of a Hessian matrix $\nabla^2\mathcal{L}_{pt}(\theta)$, is computational expensive due to its $O(n^3)$ complexity. Various methods have been proposed to approximate the inverted Hessian matrix, including Neumann series~\citep{pmlr-v108-lorraine20a},  conjugate gradients~\citep{imaml} and finite difference~\citep{pmlr-v139-idarts}. In TapWeight, we select finite difference as the approximation method, thus enabling efficient computation of best-response Jacobian matrices.

\section{Molecule Property prediction}

\subsection{Pretraining Objectives}
\label{appen:molecule_task}

We use 3 types of pretraining objectives for continued pretraining of an Imagemol model~\citep{imagemol} to enhance its performance on molecule property prediction tasks.

\paragraph{Multi-Granularity Clustering} In this pretraining objective, we first perform K-means clustering to the unlabeled training dataset of molecules using their chemical structural fingerprint. After clustering, each molecule is assigned with a pseudo-label, and the molecular encoder model is pretrained by predicting this label. Formally,

\begin{align}
    & \mathcal{L}_{mg1} = \sum_{i=1}^n \mathcal{L}(C^{100}(f_{\theta}(x_i)), y_i^{100})\\
     & \mathcal{L}_{mg2} = \sum_{i=1}^n \mathcal{L}(C^{1,000}(f_{\theta}(x_i)), y_i^{1,000})\\
     & \mathcal{L}_{mg3} = \sum_{i=1}^n \mathcal{L}(C^{10,000}(f_{\theta}(x_i)), y_i^{10,000})
\end{align}

where $f_{\theta}$ is the molecular encoder, and $C$ are task-specific fully-connected neural networks for clustering label prediction.

\paragraph{Mask-based Contrastive Learning} In this pretraining objective, we use a 16 × 16 square area to randomly mask a molecular image $x$ to generate the masked image $\hat{x}$. We then perform constrastive learning on the image pair $(x, \hat{x})$ by minimizing the distance between representations of both images to promote consistency. Formally,

\begin{align}
    & \mathcal{L}_{mcl} = \sum_{i=1}^n || f_{\theta}(x_i), f_{\theta}(\hat{x_i}) ||_2
\end{align}

where $||\cdot||$ denotes the Euclidean distance between two molecular representation generated from the encoder.

\paragraph{Jigsaw Puzzle Prediction} In this pretraining objective, we introduce 100 types of different permutations with number 1 to 100, denoted as $y^{jig}$. We also assign a label of 0 for original molecular image without any  We apply the permutation to molecular images $x$ to get permuted ones $\hat{x}$. The encoder $f_{\theta}$ is pretrained by predicting the permutation label. Formally,

\begin{align}
    & \mathcal{L}_{jpp} = \sum_{i=1}^n \mathcal{L}(C(f_{\theta}(\hat{x_i})), y_i^{jig})
\end{align}

where $C$ is a task-specific fully-connected neural network for permutation label prediction.

\subsection{Datasets}
\label{appen:molecule_data}

We use the datasets from MoleculeNet benchmark for molecule property prediction~\cite{Wu2017MoleculeNetAB}.

\paragraph{Quantum Mechanics} Qm7 and Qm9 are both molecular datasets for regression task on quantum mechanics properties of molecules. Qm7 dataset collects electronic properties of molecules determined using ab-initio density functional theory (DFT). Qm9 dataset collects geometric,  energetic, electronic and thermodynamic properties of DFT-modelled small molecules.

\paragraph{Physical Chemistry} Esol, FreeSolv and Lipophilicity (Lipo) are all datasets for regression task on physical chemistry properties of molecules. ESOL dataset collects water solubility data for common organic small molecules. FreeSolv dataset collects experimental and calculated hydration free energy of small molecules in water. Lipo dataset collects experimental results of octanol/water distribution coefficient.

\paragraph{Biophysics} Bace, HIV and MUV are all datasets for classification tasks on biophysics properties of molecules. BACE dataset collects binary label of molecular binding results for a set of inhibitors of human $\beta$-secretase 1 (BACE-1). HIV dataset collects experimentally measured abilities of a molecule to inhibit HIV replication. MUV is a subset of PubChem BioAssay by applying a refined nearest neighbor analysis, designed for validation of virtual screening techniques.

\paragraph{Physiology} BBBP, Clintox, Sider, Toxcast and Tox21 are all datasets for classification tasks on physiology properties of molecules. BBBP dataset contains binary labels of blood-brain barrier penetration (permeability) ability for molecules. ClinTox dataset consists of qualitative data of drug molecules approved by the FDA and those that have failed clinical trials for toxicity reasons. Sider is a database of marketed drugs and adverse drug reactions (ADR), grouped into 27 system organ classes. ToxCast dataset contains toxicology data for a large library of compounds based on in vitro high-throughput screening, including experiments on over 600 tasks. Tox21 dataset collects qualitative toxicity measurements of molecules on 12 biological targets, including nuclear receptors and stress response pathways.

\subsection{Baseline Methods}
\label{appen:molecule_bsl}

\paragraph{Attribute Masking} Attribute masking (AttrMask) based pretraining captures domain knowledge by learning the regularities of the
node/edge attributes distributed over graph structure~\citep{hu2020pretraining}. Inspired by BERT~\citep{devlin-etal-2019-bert}, it pretrains a graph neural network (GNN) by first masking node/edge attributes and then letting GNNs predict those attributes based on neighboring structure. 

\paragraph{Context Prediction } Context Prediction uses subgraphs to predict their surrounding graph structures~\citep{hu2020pretraining}. It pretrains a GNN so that it maps nodes appearing in similar structural contexts to nearby embeddings. Specifically, the method first encodes the context into a fixed vector using an auxiliary GNN, and then trains the GNN encoder with negative sampling.

\paragraph{GraphMVP} The Graph Multi-View Pre-training (GraphMVP) framework applies self-supervised learning (SSL) by utilizing the correspondence and consistency between 2D topological structures and 3D geometric views~\citep{liu2022graphmvp}. It introduces a novel contrastive learning loss, using the 2D and 3D representations of the same molecule as positive pairs.

\paragraph{Imagemol} ImageMol is an unsupervised pretraining deep learning framework pretrained on 10 million unlabelled drug-like, bioactive molecules, to predict molecular targets of candidate compounds~\citep{imagemol}. The ImageMol framework is designed to pretrain chemical representations from unlabelled molecular images on the basis of local and global structural characteristics of molecules from pixels.

\subsection{Hyperparameter Settings}
\label{appen:molecule_hyper}

\paragraph{Classification} We set the global learning steps to be 30,000 for MUV dataset, 20,000 for HIV dataset, 10,000 for Tox21 and Toxcast datasets, and 3,000 for all other datasets. We set the batch size in level I to be 1024, and that in level II and level III to be 64 for all datasets. We set the learning rate to be 0.02 in level I, 0.05 in level II, and that in level III to be 200 for all datasets. We set the $\gamma$ value in Equation \ref{eq:level2} to be 0.001. 

\paragraph{Regression} We set the global learning steps to be 30,000 for qm9 dataset and 10,000 for all other datasets. The batch size and $\gamma$ are the same as those in classification tasks. We set the learning rate to be 0.02 in level I, 0.001 in level II, and 1 in level III for Lipo, Esol and FreeSolv datasets. We set the learning rate to be 0.01 in level I, 0.0001 in level II, and 0.1 in level III for Qm7 and Qm9 datasets. 

\section{Natural Language Understanding}

\subsection{Pretraining Objectives}
\label{appen:nlu_task}

We use 3 types of losses for continued pretraining of an RoBERTa model~\cite{liu2019roberta} to enhance its performance on natural language understanding tasks.

\paragraph{Mask Language Modeling} This pretraining objective randomly mask some percentage of the input tokens, and then predict those masked tokens using embedding generated from the pretrained model~\citep{devlin-etal-2019-bert}. In BERT and RoBERTa, 15\% of the tokens are masked in the pretraining stage.

\paragraph{Constrastive Learning} This pretraining objective applies dropout noise to the encoder $f_{\theta}$ when taking in a sentence $x$ to get a negative sample of encoding $h'=f_{\theta}(x)$~\citep{gao2021simcse}. We use $h$ to denote those positive encodings without dropout noise. The encoder is then trained by minimizing a constrastive learning loss:

\begin{align}
    & \mathcal{L}_{cl} = -\sum_{i=1}^n \log(\frac{e^{sim(h_i, h_i') }}{\sum_{j=1}^ne^{sim(h_i, h_j')}})
\end{align}

where $sim$ is a similarity measure between two encodings. 

\paragraph{Sentence Order Prediction} This pretraining objective uses two consecutive segments from the same document as positive examples. It generates negative examples using the same two consecutive segments but with their order swapped~\citep{Lan2020ALBERT:}. The model is pretrained by predicting the label of these two types of examples. 

\subsection{Datasets}
\label{appen:nlu_data}

We use 8 datasets from GLUE benchmark in natural language understanding tasks~\citep{wang2018glue}.  

\paragraph{Single Sentence Tasks} The Corpus of Linguistic Acceptability (CoLA) contains English acceptability judgments sourced from books and journal articles on linguistic theory. The Stanford Sentiment Treebank (SST-2) features sentences from movie reviews annotated by humans for sentiment analysis.

\paragraph{Similarity and Paraphrase Tasks} The Microsoft Research Paraphrase Corpus (MRPC) is a dataset of sentence pairs extracted from online news sources, annotated by humans for semantic equivalence. The Quora Question Pairs (QQP) dataset includes question pairs from the Quora website, where the task is to determine if the questions are semantically equivalent. The Semantic Textual Similarity Benchmark (STS-B) contains sentence pairs from news headlines, video and image captions, and natural language inference datasets, with the task of predicting a human-annotated similarity score.

\paragraph{Inference Tasks} The Multi-Genre Natural Language Inference Corpus (MNLI) is a crowdsourced dataset of sentence pairs annotated for textual entailment, where the task is to predict the relationship between a premise and a hypothesis. Question-answering Natural Language Inference (QNLI) involves question-paragraph pairs, with the task of determining whether the paragraph contains the answer to the question. The Recognizing Textual Entailment (RTE) datasets consist of sentence pairs from news and Wikipedia, where the task is to predict the entailment between two sentences.

\subsection{Baseline Methods}
\label{appen:nlu_bsl}

\paragraph{RoBERTa} The Robustly Optimized BERT Pretraining (RoBERTa) paper~\citep{liu2019roberta} thoroughly evaluates the impact of key hyperparameters and training data size in BERT. RoBERTa uses the same architecture as BERT but is pretrained with an optimized strategy, leading to significant improvements in performance across various downstream tasks. The main differences between RoBERTa and BERT are: (1) training for a longer duration with larger batches and more data; (2) removing the next sentence prediction objective; (3) training on longer sequences; and (4) dynamically adjusting the masking patterns applied to the training data.

\paragraph{SimCSE} The Simple Contrastive Learning of Sentence Embeddings (SimCSE) framework includes both unsupervised and supervised approaches. In the unsupervised approach, SimCSE takes an input sentence and predicts the same sentence using a contrastive objective, where standard dropout serves as the noise. In the supervised approach, it integrates annotated pairs from natural language inference datasets into the contrastive framework, using human-labeled "entailment" pairs as positive examples and "contradiction" pairs as hard negatives.

\subsection{Hyperparameter Settings}
\label{appen:nlu_hyper}

We set the global learning steps to 20,000 for the QQP and MNLI datasets, and 10,000 for all other datasets. The batch size for level I is set to 512, while for levels II and III, it is set to 32 across all datasets. The learning rate for levels I and II is 2e-5, and for level III, it is set to 1 for all datasets. We set the $\gamma$ value in Equation \ref{eq:level2} to be 0.005. 

\end{document}

%% file: math_commands.tex
\usepackage{amsmath,amsfonts,bm}

\def\eqref#1{equation~\ref{#1}}

\def\1{\bm{1}}

\DeclareMathAlphabet{\mathsfit}{\encodingdefault}{\sfdefault}{m}{sl}
\SetMathAlphabet{\mathsfit}{bold}{\encodingdefault}{\sfdefault}{bx}{n}

\DeclareMathOperator*{\argmin}{arg\,min}